\documentclass[letterpaper]{article}

\usepackage[utf8]{inputenc}
\usepackage{amsmath,amscd,amssymb,amsthm}
\usepackage{fullpage}
\usepackage[numbers]{natbib}
\usepackage{hyperref}

\usepackage{microtype}
\usepackage{graphicx}
\usepackage{booktabs} 

\usepackage{amsmath,amscd,amssymb,amsthm}
\usepackage{verbatim}
\usepackage{thmtools}
\usepackage{thm-restate}
\usepackage{algorithm}
\usepackage{xcolor}
\usepackage{siunitx}
\usepackage{microtype}      
\usepackage{caption}
\usepackage{subcaption}

\usepackage{algpseudocode}

\newcommand{\argmin}{\mathop{\text{argmin}}}

\algblock{ParFor}{EndParFor}
\algnewcommand\algorithmicparfor{\textbf{parfor}}
\algnewcommand\algorithmicpardo{\textbf{do}}
\algnewcommand\algorithmicendparfor{\textbf{end\ parfor}}
\algrenewtext{ParFor}[1]{\algorithmicparfor\ #1\ \algorithmicpardo}
\algrenewtext{EndParFor}{\algorithmicendparfor}

\newcommand{\bw}{\vec{w}}

\newcommand{\bh}{\vec{h}}

\newcommand{\grad}[1]{\nabla_{#1}}
\newcommand{\gradgrad}[1]{\nabla^2_{#1}}

\renewcommand{\O}{\mathcal{O}}

\usepackage{hyperref}

\title{ALX: Large Scale Matrix Factorization on TPUs\vspace{1em}}
\author{Harsh Mehta \qquad Steffen Rendle \qquad Walid Krichene \qquad Li Zhang\\
Google Research\\
{\tt\small \{harshm, srendle, walidk, liqzhang\}@google.com}
}
\date{}
\begin{document}

\maketitle

\begin{abstract}
We present ALX, an open-source library for distributed matrix factorization using Alternating Least Squares, written in JAX. Our design allows for efficient use of the TPU architecture and scales well to matrix factorization problems of O(B) rows/columns by scaling the number of available TPU cores. In order to spur future research on large scale matrix factorization methods and to illustrate the scalability properties of our own implementation, we also built a real world web link prediction dataset called WebGraph. This dataset can be easily modeled as a matrix factorization problem. We created several variants of this dataset based on locality and sparsity properties of sub-graphs. The largest variant of WebGraph has around 365M nodes and training a single epoch finishes in about 20 minutes with 256 TPU cores. We include speed and performance numbers of ALX on all variants of WebGraph. Both the framework code \footnote{Code: \url{https://github.com/google-research/google-research/tree/master/alx}} and the dataset \footnote{WebGraph Dataset: \url{https://www.tensorflow.org/datasets/catalog/web_graph}} are open-sourced.
\end{abstract}

\section{Introduction}

Matrix factorization is one of the core techniques in the field of recommender system and for analyzing graphs.
Despite the rise of more complex neural models, matrix factorization can still achieve competitive performance  in recommender benchmarks (e.g.,~\cite{rendle:baselines,rendle:ialsbenchmarks}).
Alternating least squares (ALS), and especially its implicit variation~\cite{hu:ials}, is a fundamental algorithm to learn the parameters of matrix factorization.
ALS is known for its high efficiency, scaling linearly in both the number of rows, columns and non-zeros.
Hence, this algorithm is very well suited for large scale problems.

In a somewhat orthogonal direction, recent success of deep learning has spurred a new wave of research and progress on hardware accelerators. As the training set and model sizes grow, novel strategies for distributing the computation and model weights have been explored. In order to make these efforts cost effective, domain specific hardware acceleration is considered. One such notable hardware accelerator is Google's Tensor Processing Unit (TPU). A full pod of the current generation TPU v3 can provide 100+ petaflops of compute and 32 TiB of high-bandwidth memory, distributed across 2048 individual devices which are connected in a 2D toroidal mesh network over high-speed interconnects \cite{tpu_types}.

While most of the distributed implementations of matrix factorization we know of leverage off-the-shelf CPU devices \cite{zhuang2013,yun2014nomad,schelter2014factorbird,facebook2015recommending}, it is tempting to ask whether a high performance implementation can be devised on a large scale cluster of hardware accelerators. Our optimism stems from the facts that (1) A TPU pod has enough distributed memory to store very big sharded embedding tables; (2) TPUs are devised for workloads which can benefit from data parallelism, this is useful for solving a large batch of system of linear equations, a core operation for Alternating Least Squares; (3) TPU chips are interconnected directly with dedicated, high bandwidth and low latency interconnects. This makes gather and scatter operations over a large distributed embedding table stored in TPU memory feasible. And finally, (4) since any node failure can lead to a halt in training process, traditional ML workloads require a highly reliable distributed setup, a requirement which a cluster of TPUs can fulfill. This property enables sharding a large embedding table possible over all available devices without worrying much about replication and fault tolerance issues. 

To exploit aforementioned appealing properties of the TPU architecture, in this work, we designed and analyzed an implementation of matrix factorization method using ALS which shows strong performance both in terms of speed and scalability. We start with discussing the design choices we made in order to come up with the architecture. Then we introduce the WebGraph dataset, a large scale web link prediction dataset. Our main motivation for creating this dataset was to illustrate the scaling properties of ALX. In addition, due to increase in size of most real world problems, we feel that scalability is a crucial axis to evaluate any future advancement in matrix factorization and we hope that this dataset can be useful to evaluate scaling properties of all future generations of matrix factorization methods. Evaluation results of ALX is provided on all variants of WebGraph dataset with scaling analysis to demonstrate high parallel efficiency of the proposed implementation. Finally, we discuss a few pertinent issues related to precision and linear system solvers. 

\begin{figure*}
    \centering
\includegraphics[width=.7\textwidth]{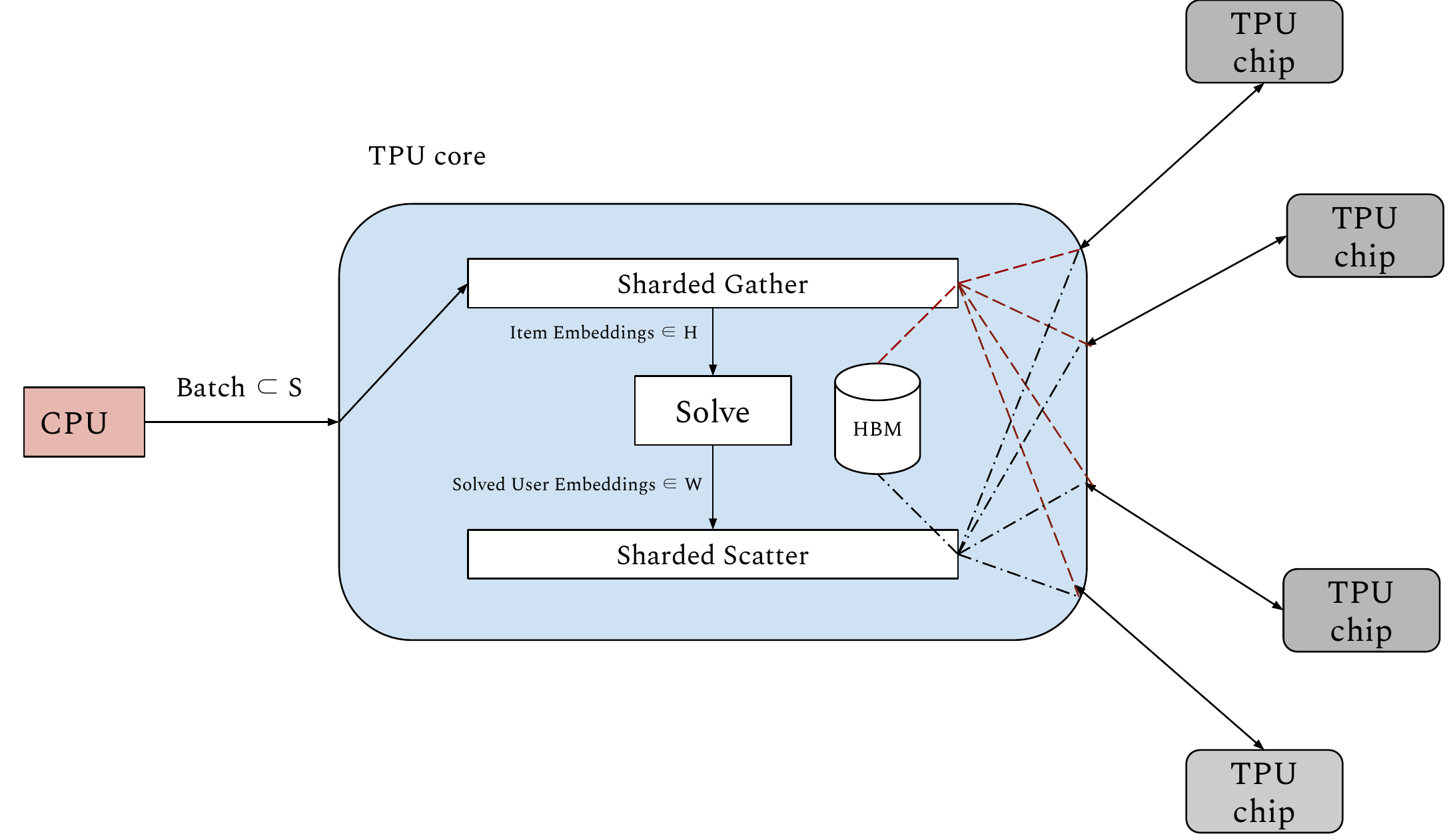}
    \caption{The flow of data and computation through ALX framework on TPU devices. Each TPU core performs identical Sharded Gather, Solve and Sharded Scatter stages for its batch of data in SPMD fashion.}\label{fig:tpu}
\end{figure*}

\section{Related Work}
Existing large-scale implementations of matrix factorization are either based on stochastic gradient descent (SGD) or Alternating Least Squares (ALS).

The majority of implementations are based on SGD, and have adopted different distribution strategies that we briefly overview. LibMF~\cite{zhuang2013} provides a multi-core, single-machine implementation, which limits its scalability. \cite{yun2014nomad,schelter2014factorbird} consider a distributed implementation which uses a parameter server architecture where the model parameters are stored on highly available servers, and the training data are processed in parallel by workers. The main bottleneck in such approaches is their high communication cost: when processing a (user, item) training example, the worker fetches the corresponding embeddings, so that one epoch of training requires sending $O(|S| d)$ parameters over the network, where $S$ is the training data and $d$ is the embedding dimension. To alleviate this communication bottleneck, Facebook~\cite{facebook2015recommending} subsequently proposed a distribution scheme wherein the embeddings are communicated in a circular fashion in rounds: during each round, a worker obtains a subset of the item embeddings, and processes updates involving those items. The communication cost becomes $O(ndw)$ where $n$ is the number of items, and $w$ is the number of workers, which can be potentially beneficial when $nw \ll |S|$, but ultimately has limited scalability due to the linear dependence on number of workers.

While simpler to implement at scale, SGD-based methods are typically not suitable to problems with implicit feedback, where the loss function is formulated as a sum over all user-item pairs, and not just over the observed pairs. A recent attempt employs efficient gramian estimation in order to scale SGD based solution \cite{gravity}. Still, ALS is known to converge faster than SGD for such implicit feedback problems~\cite{hu:ials}.

Although ALS is harder to implement at scale, a few attempts were made. Spark MLlib~\cite{spark2014scalable} opts for partial model replication, wherein if a worker processes a shard of training data, it has a copy of all embeddings involved in the solution for that shard, which in the worst case amounts to full model replication. This approach is suitable to small and dense problems in which model replication is feasible, but does not scale to very large models. GraphLab~\cite{low2012distributed} uses a parameter server approach and suffers from the same communication bottleneck described above. The circular communication scheme described in~\cite{facebook2015recommending} is also applicable to ALS, where partial Gramian matrices are communicated between workers in a circular fashion. This reduces the communication bottleneck to $O(wnd^2)$ but the quadratic dependency on $d$ make this less scalable than the SGD variant. All the aforementioned implementations are CPU-based. To take advantage of hardware acceleration, \cite{tan2016cuMF} propose cuMF, a single-machine, memory-optimized multi-GPU implementation that scales to relatively large problems (up to $10^{11}$ model parameters), further extended in~\cite{tan2018matrix} to allow approximate computation via a conjugate gradient solver. \cite{tan2016cuMF,tan2018matrix} exploit GPU memory hierarchy and model parallelism across GPUs in order to produce a highly performant implementation of ALS. As cuMF exploits unique properties of the GPU hardware, ALX overcomes various challenges and exploits unique properties of TPUs. This allows employing up to a full pod of 2048 TPUs for very large scale matrix factorization problems.

\section{Problem Setting}\label{sec:als}
Before introducing the architecture of ALX, which is optimized for hardware accelerators, we present the problem setting and generic flow of operations for an end to end matrix factorization using Alternating Least Squares.


Matrix factorization can model the relationship of two categorical variables.
For illustration, let one of the variable be a user $U=\{u_1, u_2, \ldots u_m\}$ and the other one items $I=\{i_1, i_2,\ldots, i_n\}$.
We are interested in the relationship $y$ between these two variables, $y : U \times I \rightarrow \mathbb{R}$.
An example would be a recommender system where $y(u,i)$ could expresses how much user $u$ likes item $i$.
This relationship can equivalently be seen as a matrix $Y \in \mathbb{R}^{U \times I}$ where $y_{u,i} = y(u,i)$.

Matrix factorization models the relationship $y$ by embedding each user and item into a $d$ dimensional embedding space and the estimated relationship $\hat{y}: U \times I \rightarrow \mathbb{R}$ is defined as the dot product between the embeddings:
\begin{align}
    \hat{y}(u,i) = \langle \bw_u, \bh_i \rangle
\end{align}
where $W \in \mathbb{R}^{U \times d}$ is the embedding matrix of users and $H \in \mathbb{R}^{I \times d}$ the embedding matrix of the items.

The model parameters of this matrix are learned based on a training set $S \subseteq U \times I \times \mathbb{R}$, where a tuple $(u,i,y) \in S$ indicates that the user-item pair $(u,i)$ has label $y$.
With this training set we can define the training objective as:
\begin{align}
    \argmin_{W,H} \sum_{(u,i,y) \in S} (y-\hat{y}(u,i))^2 + \lambda ||W||_F^2+ \lambda ||H||_F^2
\end{align}
where $\lambda \in \mathbb{R}^+_0$ is a L2 regularization hyper parameter.

In many applications, the information about which user-item pairs are in $S$ carries information.
For example, which items have not been purchased by a user can carry important information about a user's preference.
Thus it is common~\cite{hu:ials} to add also all possible pairs as weakly negative feedback into the objective:
\begin{align}
    \argmin_{W,H} \sum_{(u,i,y) \in S} (y-\hat{y}(u,i))^2 + \alpha \sum_{u \in U} \sum_{i \in I} \hat{y}(u,i)^2 + \lambda ||W||_F^2+ \lambda ||H||_F^2
\end{align}
Here $\alpha \in \mathbb{R}^+_0$ is a hyper parameter that controls the strength of the weakly negative pairs.
The additional term is especially important for settings where only positive feedback has been observed (e.g., user $u$ has bought product $i$).
In such cases, the weakly negative feedback is crucial for learning a meaningful model.

\cite{hu:ials} have proposed to solve this objective by alternating least squares.
When $H$ is fixed, $W$ has a closed form solution.
The optimal choice for each user embedding $\bw_u$ is the solution of a linear regression problem and its solution is:
\begin{align}
    \bw_{u^*} \leftarrow \left(\sum_{(u^*,i,y) \in S} \bh_i \otimes \bh_i + \alpha \sum_{i \in I} \bh_i \otimes \bh_i  + \lambda I\right)^{-1} {\sum_{(u^*,i,y) \in S} \bh_i y}. \label{eq:ialsupdate}
\end{align}
A key observation of \cite{hu:ials} is that the term $\sum_{i \in I} \bh_i \otimes \bh_i = H^t\,H$, the Gramian of $H$, is independent of the user $u$ and can be precomputed.
After the user embeddings have been optimized, a similar step can be made for optimizing the item embeddings while holding the user embeddings constant.
The algorithm alternates between the user and item pass until convergence.
The overall  algorithm is sketched in Algorithm~\ref{alg:ials}.
A full epoch of this algorithm has a computational complexity of $\O(d^2 |S| + d^3\,(|U| + |I|))$.
Empirically, the algorithm converges after a few alternating steps, for example 16 epochs is usually sufficient.

\begin{algorithm}[t]
\caption{ALS algorithm for implicit data \cite{hu:ials}\label{alg:ials}}
\begin{algorithmic}[1]
    \For {$t \in \{1,\ldots,T\}$}
        \State $G^I \leftarrow H^t\, H$ \Comment{$\O(|I| d^2)$}
        \For{$u^* \in U$} \label{ln:ials_user_loop}
            \State $\grad{\bw_{u^*}} \leftarrow 0$ 
            \State $\gradgrad{\bw_{u^*}} \leftarrow \alpha  G + \lambda$ \Comment{$\O(d^2)$}
            \For{$(u^*,i,y) \in S$} 
                \State $\grad{\bw_{u^*}} \leftarrow \grad{\bw_{u^*}}  + y \bh_i$ \Comment{$\O(d)$}
                \State $\gradgrad{\bw_{u^*}} \leftarrow \gradgrad{\bw_{u^*}}  +  \bh_{i} \otimes \bh_{i}$ \Comment{$\O(d^2)$}
            \EndFor
            \State $\bw_{u^*} \leftarrow (\gradgrad{\bw_{u^*}})^{-1} \grad{\bw_{u^*}}$\Comment{$\O(d^{3})$} 
        \EndFor
        \State{Perform a similar pass over the item side}
    \EndFor
\end{algorithmic}
\end{algorithm}



\section{ALX}\label{sec:architecture}
\begin{algorithm}[t]
\caption{ALX algorithm for TPU core $\mu$\label{alg:alx}}
\begin{algorithmic}[1]
    \State Initialize TPU core $\mu \in M$
    \State Initialize User embedding shard $W_\mu$
    \State Initialize Item embedding shard $H_\mu$
     \For {$t \in \{1,\ldots,T\}$}
    \State $G^I_\mu \leftarrow H^T_\mu H_\mu$ 
    \State $G^I \leftarrow \sum_{\mu' \in M} G^I_{\mu'}$ \Comment{Collective: all reduce sum of Gramians}
    \For{$U_k(\mu) \subseteq U$} \Comment{Iterate over batches/ subsets of users}
        \State $I_k(\mu) \leftarrow \{ i : (u', i, \cdot) \in S, u' \in U_k(\mu) \}$
        \State $H_{I_k(\mu)} \leftarrow \text{sharded\_gather}(H, I_k(\mu))$ \Comment{Collective: gather item embeddings from all the shards}
         \ParFor{$u^* \in U_k(\mu)$}
            \State $\grad{\bw_{u^*}} \leftarrow 0$ 
            \State $\gradgrad{\bw_{u^*}} \leftarrow \alpha  G^I + \lambda$ \Comment{$\O(d^2)$}
            \ParFor{$(u^*,i,y) \in S$} 
                \State $\grad{\bw_{u^*}} \leftarrow \grad{\bw_{u^*}}  + y \bh_i$ \Comment{$\O(d)$}
                \State $\gradgrad{\bw_{u^*}} \leftarrow \gradgrad{\bw_{u^*}}  +  \bh_{i} \otimes \bh_{i}$ \Comment{$\O(d^2)$}
            \EndParFor
            \State $\bw_{u^*} \leftarrow (\gradgrad{\bw_{u^*}})^{-1} \grad{\bw_{u^*}}$\Comment{$\O(d^{3})$}
        \EndParFor
        \State $W \leftarrow \text{sharded\_scatter}(W, W_{U_k(\mu)})$ \Comment{Collective: scatter user embeddings across all the shards}
     \EndFor
        \State{Perform a similar pass over the item side}
    \EndFor
\end{algorithmic}
\end{algorithm}

Distributed machine learning architectures come in all shapes and sizes. In our context, we need a distributed storage to keep the embedding tables ($H$ and $W$) with the ability to do fast gather and scatter operations on them (see line 6-9 in Algorithm~\ref{alg:ials}). As illustrated in Figure \ref{fig:tpu}, in this section we describe how the embedding tables are sharded, the flow of data and computation through one epoch and a few salient challenges we faced. The design choices we made are laid out concretely as Algorithm \ref{alg:alx}. Before we dive into ALX architecture, we provide a brief overview of the TPU system architecture. Finally, we also present a few implementation details we found to be crucial.

\subsection{TPU architecture}
Each host CPU is connected to 4 TPU chips in a single board. Each chip has 2 cores. Each TPU core has scalar, vector, and matrix multiplication units (MXUs). The MXUs provide the bulk of the compute power in a TPU chip. Each MXU is capable of performing 16K multiply-accumulate operations in each cycle at reduced bfloat16 precision \cite{bfloat16}. The inputs and outputs can have float32 precision but the multiply-accumulate (MAC) operations on MXU are performed with bfloat16. Note that a single float32 number can be decomposed into multiple bfloat16 numbers and float32 MAC operations can be achieved but at reduced performance.

Typically, system resources on a TPU core vary by version. We employ TPU v3 for all out experiments. Each TPU v3 core has 16 GiB of high-bandwidth memory (HBM). A TPU v3 pod can connect a total number of 2048 of such cores, which altogether makes 32 TiB of high-bandwidth distributed memory available for use. Moreover, in a TPU pod, chips are connected through dedicated high-speed, low latency interconnects. They are connected in a 2-D toroidal mesh with each chip connected to its four neighbors such that the communication takes place in four directions. These interconnects are dedicated to chip-chip communication and  bypasss the CPU networking resources entirely.

We exploit both the size of distributed memory available and dedicated low-latency interconnects in order to efficiently shard very large embedding tables and perform efficient distributed gather and scatter operations. Note that memory usage of a TPU system can be determined at compile time. ALX program code is first converted by JAX into an intermediate representation (IR) which is then fed to the Accelerated Linear Algebra (XLA) compiler. XLA converts it into Low Level Optimizer (LLO) code which can be thought of as aseembly code for the TPU system. XLA requires that the shapes of all the tensors should be known at compile time in order for it to perform further compiler level optimizations. This can be a hard constraint to circumvent at first, especially for usecases like sparse matrix factorization. We discuss this at length in Section \ref{subsec:dense_batching}.

\subsection{Algorithm Description}
\label{subsec:gather}
\begin{figure*}
    \centering
\includegraphics[width=.37\textwidth]{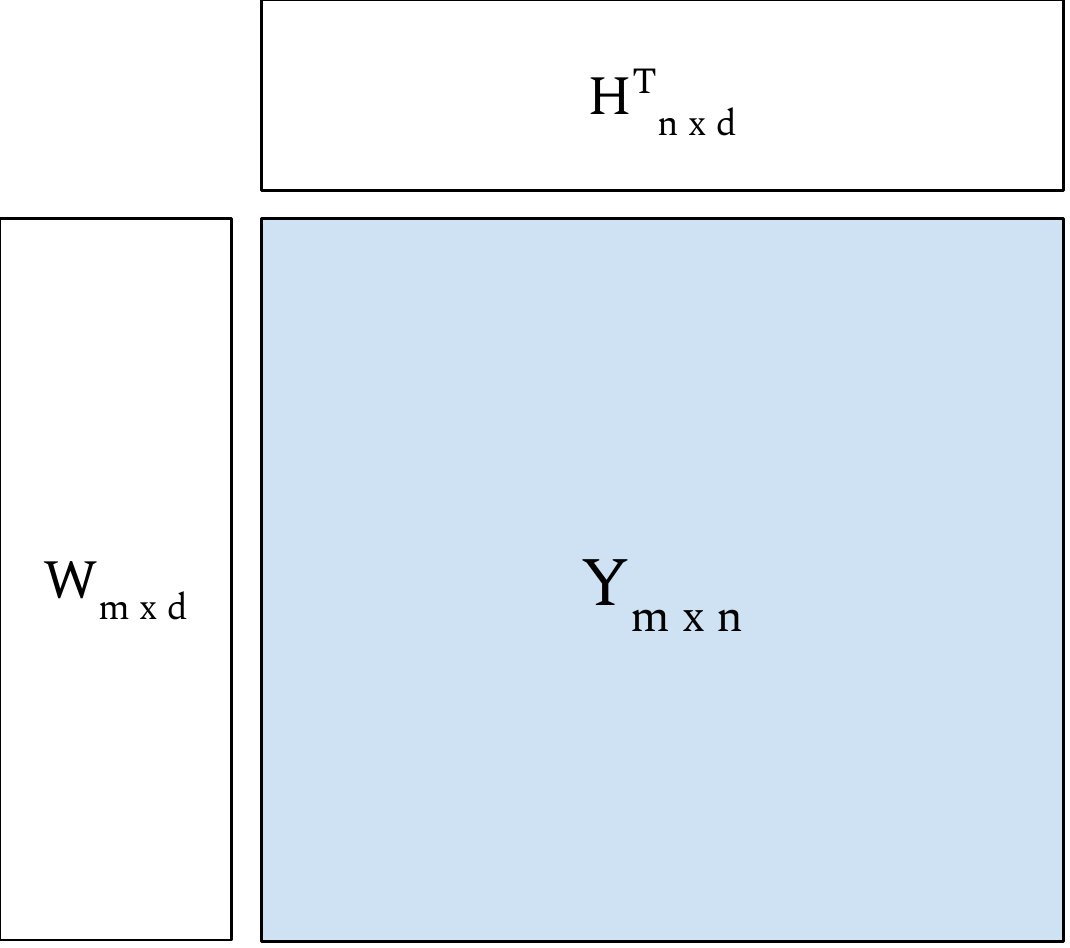}
\hspace{1.0 cm}
\includegraphics[width=.5\textwidth]{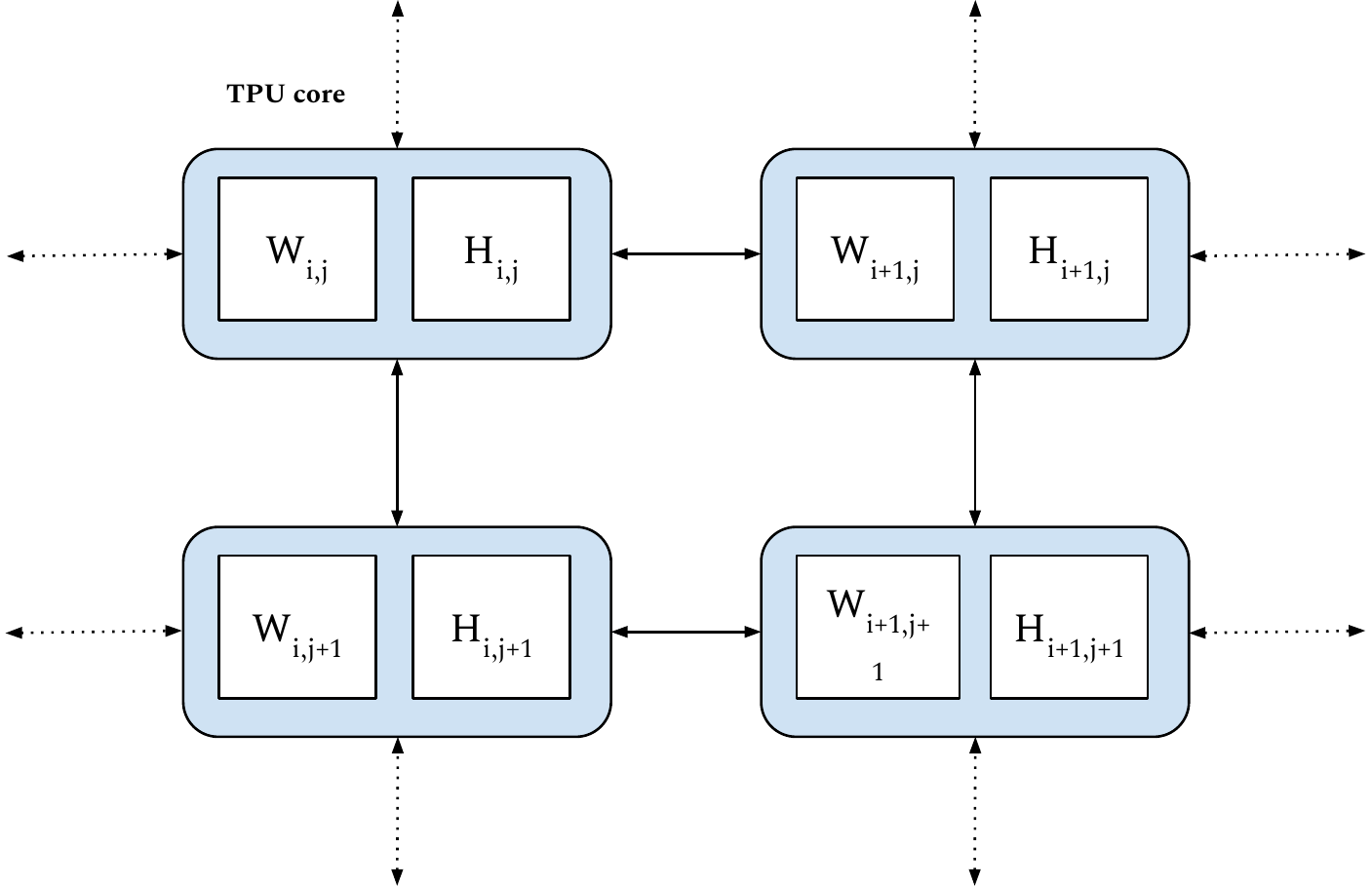}
    \caption{Uniform sharding of both embedding tables (W and H) across TPU cores. We define $m=|U|$ and $n=|I|$ from notation in Section \ref{sec:als}. Indices $i$ and $j$ refer to the coordinates of a single TPU core relative to the rest of the available TPUs.}\label{fig:shard}
\end{figure*}

In order to fully utilize the available TPU memory, we uniformly shard both user and item embedding tables (W and H) across TPU cores. As shown in Figure \ref{fig:shard}, each TPU core $\mu$  holds a fraction $W_\mu$ and $H_\mu$ of both $W$ and $H$. 

As shown in Fig \ref{fig:tpu}, a batch of data consisting of several user histories ($I_k(\mu)$) are continuously fed from the host CPU to TPU devices connected to that host. In a pod configuration, multiple hosts are employed in this process, each host connected to 8 TPU cores. The flow of computation is entirely identical and parallelized across distinct batches passed to TPU devices. This strategy is often referred to as Single-Program-Multiple-Data (SPMD) technique where the same computation is run on different input data in parallel on different devices.

\paragraph{sharded\_gather} Once the batch arrives in the memory of TPU core $\mu$, the first order of business is to gather embeddings ($\bh_i$) for every item id in those histories. Since the embedding tables $W$ and $H$ are sharded, its likely that the embedding for a certain item id is not present in the item embedding shard that a particular device holds ($H_{\mu}$ from Algorithm \ref{alg:alx}). Thus, we first run a collective \emph{all gather} operation on batch of user histories to collect user histories from all the TPU devices locally. Then each device attempts to gathers embeddings for all the user histories collected in the previous step from its own shard of the embedding table. Note that at this point, some embeddings will be invalid since they were not present in $H_{\mu}$. Since we know the bounds of the shards of embedding tables on each device, we can zero out the invalid embeddings.

At this stage, each device contains embeddings for all the user histories to be processed by all the devices. Since for any item id $i$, exactly one device holds the embedding of that item in its shard, only that device will be able to gather a non-zero embedding for that item. We exploit this constraint and run \emph{all reduce sum} collective operation on user history embedding tensor. This completes our sharded gather stage, line 9 in Algorithm \ref{alg:alx}. At the end, each device contains valid embeddings for all the items in its own batch of user histories ($H_{I_k(\mu)}$).

\paragraph{Solve} Once the item embeddings are present locally in each TPU device, they can be used to accumulate the sufficient statistics $\grad{\bw_{u^*}}$ and $\gradgrad{\bw_{u^*}}$, which are further used to solve a batch of linear system of equations as mentioned in Algorithm \ref{alg:alx} lines 10-18. There are several alternatives for solving a linear system of equations (line 17), we implement and analyze their performance in Section \ref{subsec:linear_solvers}.

\paragraph{sharded\_scatter} Once the TPU device solves for its batch of user embeddings ($\bw_{u^*}$), they need to be updated in the corresponding shards of the user embedding table $W$. To perform this sharded scatter of user embeddings, we employ a strategy similar to sharded gather. More concretely, we start with running \emph{all gather} collective on solved user embedding tensor to collect user embeddings from every device. Then the user embeddings which are out of bounds for this device's shard of the user embedding table are zeroed out. Lastly, user embeddings are added to corresponding indices in the user embedding table.

\paragraph{Gramians} Since the embedding tables $W$ and $H$ are sharded and can be big enough to not fit in any single core's memory, it is crucial that the gramian calculation can be decomposed across multiple cores as well. As shown in lines 5-6 in Algorithm \ref{alg:alx}, this is indeed possible by first calculating local gramian $G^I_\mu$ and subsequently performing an \emph{all reduce sum} across all the cores in order to obtain the global gramian $G^I$.

\paragraph{Complexity Analysis}

Next, we analyse the properties of the ALX algorithm (Alg.~\ref{alg:alx}).
The computational costs for computing the sufficient statistics $\grad{\bw_{u^*}}$ and $\gradgrad{\bw_{u^*}}$ is $\O(|S|\,d^2)$ and for solving them $\O((|U|+|I|)d^3)$.
Both tasks are distributed over $M$ cores.
The costs per core are on average $\O(\frac{1}{M}(|S|\,d^2 + (|U|+|I|)d^3))$, which implies that theoretically a linear speedup for these steps is possible compared to the non-distributed ALS algorithm (Alg.~\ref{alg:ials}).

Additionally, ALX needs to collect the embeddings $H_{I_k(\mu)}$ for computing the sufficient statistics from other cores. Since each core process one batch, number of batches being processed scales with number of available cores.  
The gather and scatter algorithms that we describe, collect all embeddings for a batch in all cores, this means in total $\O(|S|\,d\,M)$ bytes are transferred in one epoch over all machines or $\O(|S|\,d)$ per machine. Thus ALX trades off this additional network overhead for a linear data parallelism. For a single core though, this step has a constant runtime, and does not get worse with more machines.

As long as the problem (\#machines and dataset size) is bounded by the computational steps, the ALX algorithm should show linear scaling behavior that will tail off as soon as the collection of embeddings becomes the bottleneck.
We study these effects in more detail empirically in Section~\ref{subsec:empirical_scaling}.

\paragraph{Alternatives}
Note that, as mentioned in Algorithm \ref{alg:ials}, we just need $\grad{\bw_{u^*}}$ and $\gradgrad{\bw_{u^*}}$ to solve for a user embedding and not the whole list of item embeddings corresponding to that user. Thus, an alternative is to build these statistics first without any communication across TPU devices, using only the item embeddings obtained from the local shard of the embedding table on a TPU device. Once every device has partial sums of these statistics, the full statistics can be obtained by running an \emph{all reduce sum} across devices. Note that this shifts the cost of communication across TPU devices from $\O(d|S|)$ to $\O(d^2|U|)$. In our experience, this alternative performed worse in terms of running time on almost every dataset we tried.

\subsection{Dense Batching}
\label{subsec:dense_batching}
\begin{figure*}
    \centering
\includegraphics[width=.8\textwidth]{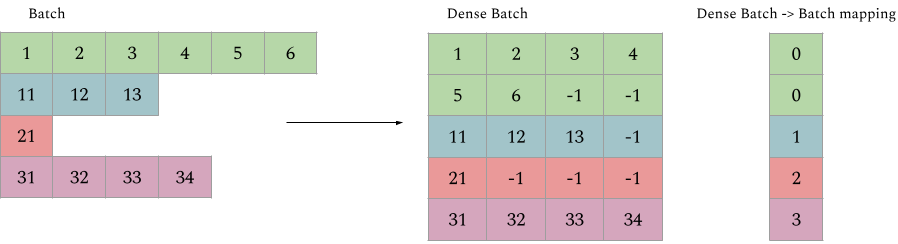}
    \caption{Illustrating example of how sparse batches are densified for it to be XLA compatible.}\label{fig:dense}
\end{figure*}
Since XLA compiler re-compiles a computation graph just in time as the shape of the input tensors change, TPUs currently are not suitable for sparse matrix representation and operations. It requires that the shapes for all the dense tensors are fixed at compile time for optimal performance. This is a major challenge in our context since the length of the input rows can be wildly different across examples. If we pad the rows naively to maximum sequence length of any example in the dataset, it can lead to a very inefficient use of resources at best since typically distribution of row lengths has a long tail.

To circumvent this crucial constraint, we employ a strategy we call Dense Batching. As shown in Figure \ref{fig:dense}, we break each row in the input sparse batch into multiple rows in the dense batch and keep an additional data structure in order to keep track of which rows in the dense batch belong to the same row in the sparse batch. If the row length of the dense batch is too big for the input example, we pad the rest of the sequence like before. The benefit of breaking the input row into multiple rows in dense batch is that if we choose a reasonable row length for the dense batch, we can avoid most of the waste incurred due to padding. On the other hand, if the row length is too small, we incur proportional cost of keeping the mapping between dense and sparse batch around. In our experience, dense row length of 8 or 16 works quite well.

\subsection{Precision}
\begin{figure*}
    \centering
\includegraphics[width=.47\textwidth]{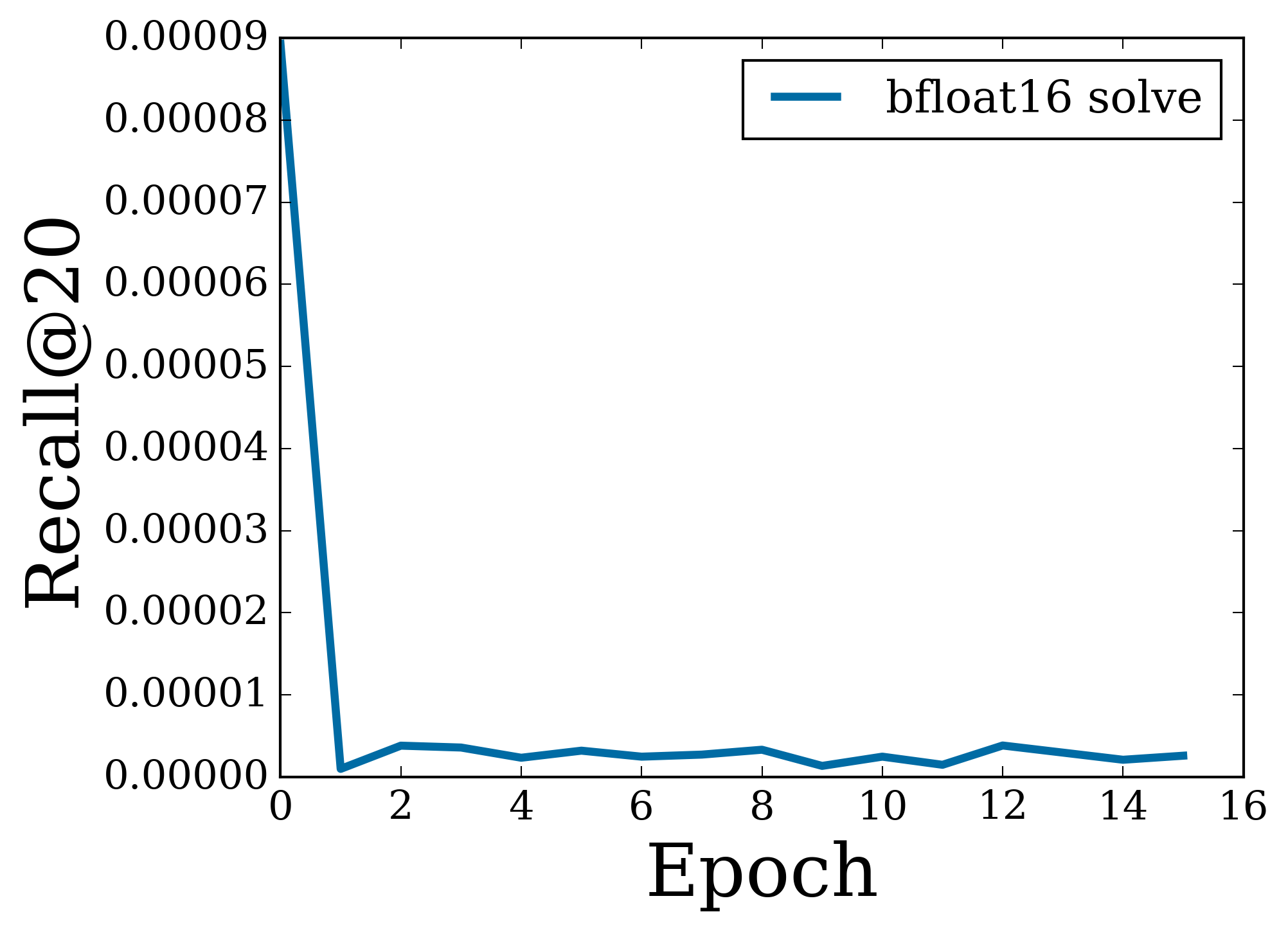}
\hspace{1.0 cm}
\includegraphics[width=.45\textwidth]{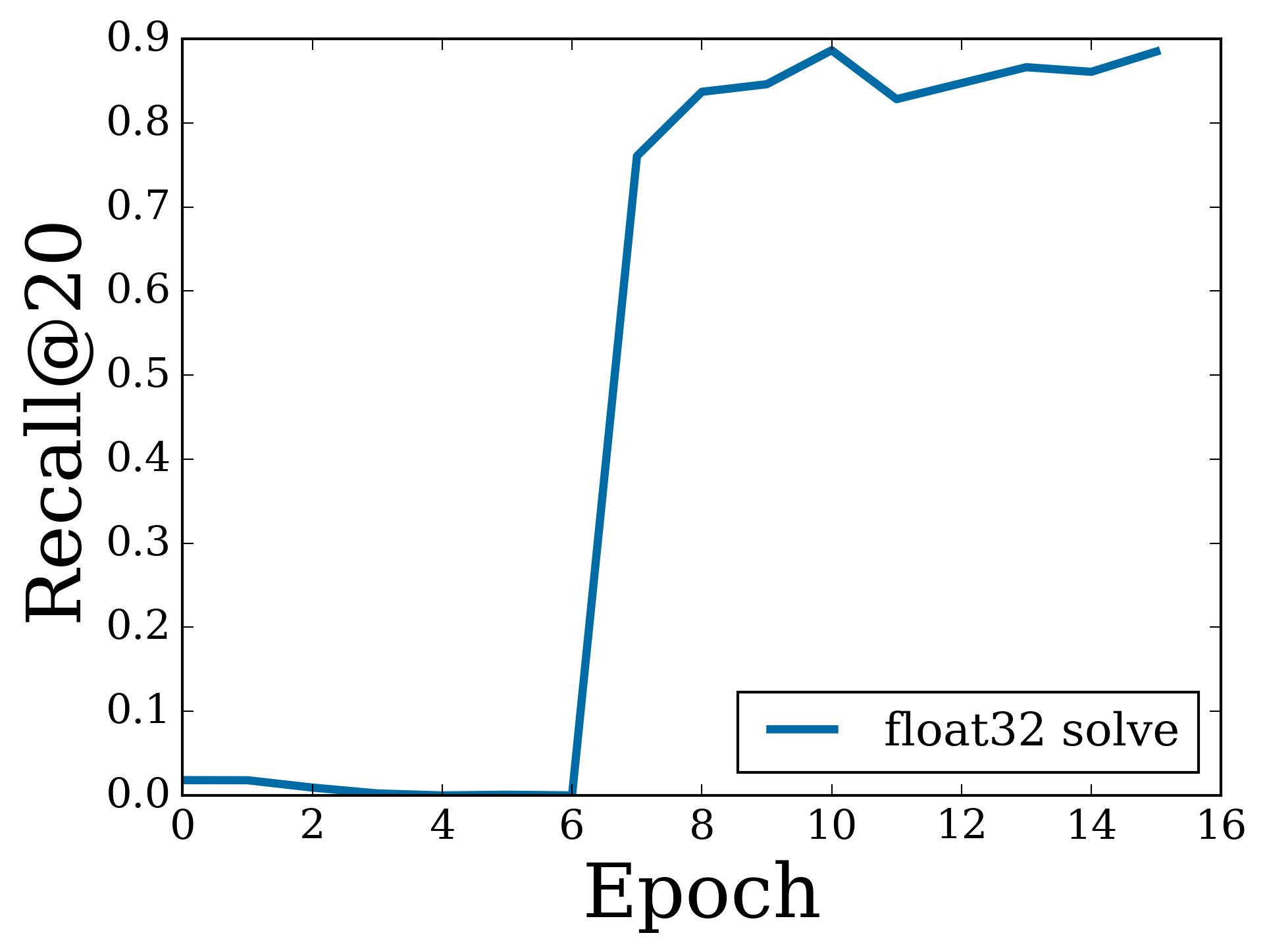}
    \caption{Comparison of eval metrics when using bfloat16 numerics vs float32. (a) shows the unrecoverable collapse in metrics when using bfloat16.}\label{fig:precision}
\end{figure*}
Since modern machine learning use cases are more tolerant to lower precision arithmetic without degrading quality of the models, accelerators are typically designed to exploit it in order to achieve impressive speed ups and memory savings. For example, TPUs natively support bfloat16 \cite{bfloat16} floating point numerics which reduces the size of data and weights in memory and allows for faster training.

In our context, as shown in Figure \ref{fig:precision}, we have found that using bfloat16 numerics naively can lead to collapses in middle of the training procedure, especially when the regularization constant is not high enough. Switching from bfloat16 to float32 entirely would result in doubling the memory requirements for storing the embedding tables and communication cost for performing sharded gather and scatter.

We have found that it is sufficient to keep both the embedding tables in bfloat16 precision, cast the input tensors for solving the linear system of equations to float32 and casting the solution back to bfloat16. This avoids incurring both the extra memory cost of storing the embeddings in higher precision and also communicating them for sharded operations.

\subsection{Linear solvers}
\label{subsec:linear_solvers}
\begin{figure*}
    \centering
\includegraphics[width=.6\textwidth]{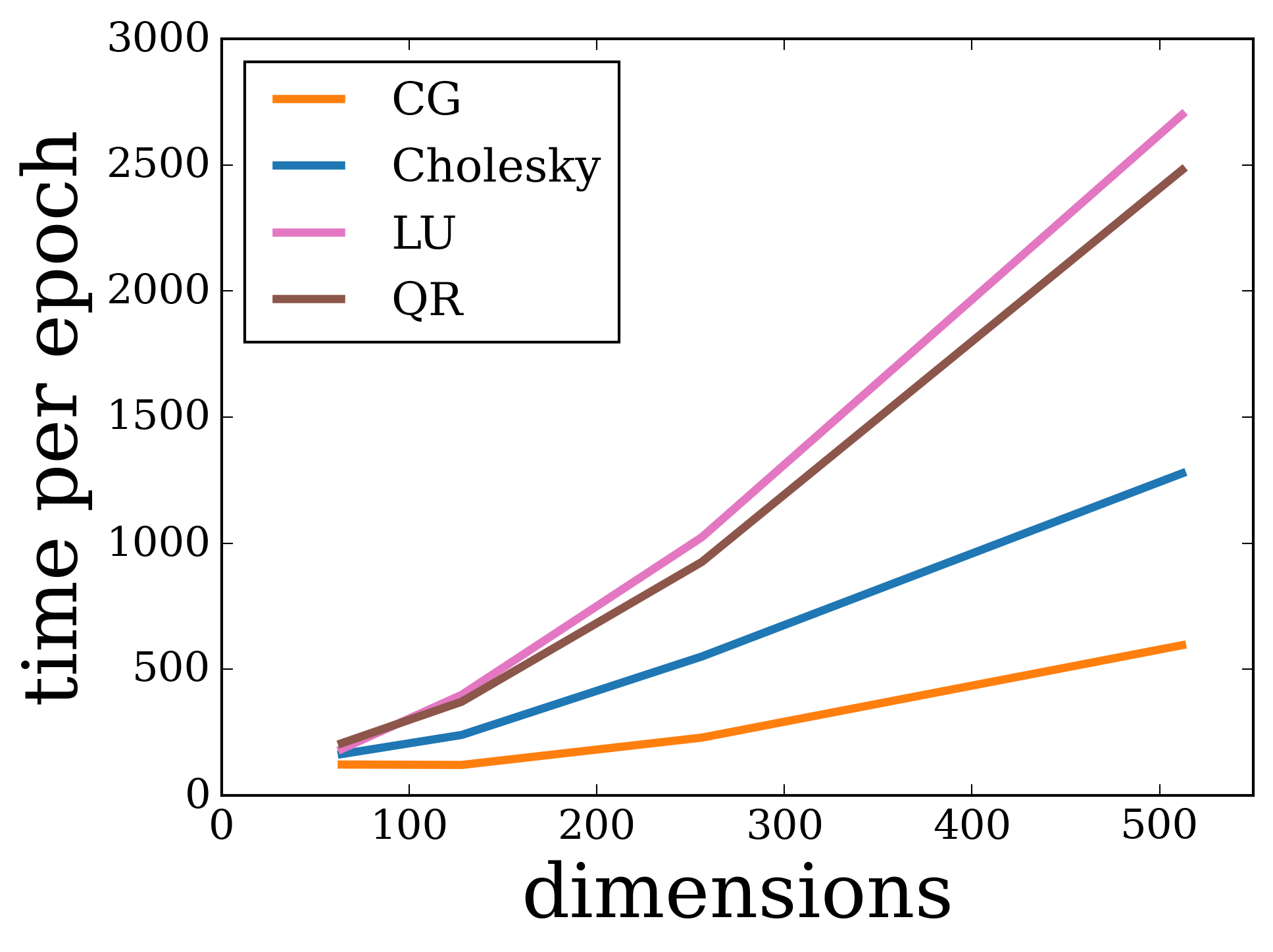}
    \caption{Comparison of training time per epoch (in seconds), of plausible alternatives of linear system solver on TPU.}\label{fig:linear_solver}
\end{figure*}
As emphasized in Algorithm \ref{alg:ials}, solving the linear regression problem has the highest computational complexity (with respect to $d$) among all the steps. Thus it is crucial to analyze possible alternative for efficiency reasons.

We compare 4 different linear regression solvers commonly used in practice; (1) LU 
(2) QR (3) Cholesky and (4) Conjugate Gradients. As shown in Figure \ref{fig:linear_solver}, on TPU, Conjugate Gradients scales most favorably with number of embedding dimensions. Since the matrix multiply unit, called the MXU, in the TPU provides for most of the parallelism, it is comforting to see that, Conjugate Gradients (an iterative algorithm) is able to utilize the MXU much more effectively than other methods which rely on operations which are hard to cast into simple matrix multiplies (e.g. LU's reliance on pivoting).


\subsection{Top-K}
Our presentation so far focused on training the model parameters. However, besides training we would like to also evaluate the solution. A typical evaluation use case is to compute the nearest neighbors of an user, e.g., to obtain the "best" k items for this user. This Top-K retreival can be slow on TPUs since it relies on comparing a value with all the other values in a list, which has $\O(n\log n)$ worst case complexity. Fortunately, this operation is needed only at either evaluation or inference time, in order to either compare or materialize results. For cases when this operation becomes painfully slow, we recommend employing a fast but approximate nearest neighbors method. Recent Maximum Inner Product Search (MIPS) algorithms can find top $k$ neighbors in running time which can scale sub-linearly with the number of items \cite{Ram2012MaximumIS,binary_codes,asymmetric_hash_topk}.   

\section{WebGraph}\label{sec:webgraph}
As emphasized earlier, our main goal is to create a highly scalable and efficient implementation of large matrix factorization. In order to perform experiments at the scale we desire and concretely understand the scaling properties of ALX, we created a large scale link prediction dataset leveraging already open Common Crawl data scraped from the web. Common Crawl has proven to be an invaluable resource for NLP tasks \cite{buck-etal-2014-n,smith-etal-2013-dirt,raffel2020exploring}. In our context, we ignore the textual information and focus on the web links. 

Common Crawl keeps a rolling, publicly available web archive of text and non-text data crawled and scraped from HTML files. Unfortunately, the link graph data can be very sparse and needs to be further processed in order to factorize it. We call the processed dataset ``WebGraph". Also, in order to increase the value of this resource, we created 6 different version of WebGraph, each varying in the sparsity pattern and locale. We took the following processing steps, in order:

\begin{itemize}
    \item We started with WAT files from June 2021 crawl.
    \item Since the outlinks in HTTP-Response-Metadata are stored as relative paths, we convert them to absolute paths using urllib after validating each link.
    \item To study locale-specific graphs, we further filter based on 2 top level domains: `de' and `in', each producing a graph with an order of magnitude less number of nodes.
    \item These graphs can still have arbitrary sparsity patterns and dangling links. Thus we further filter the nodes in each graph to have minimum of K $\in [10, 50]$  inlinks and outlinks. Note that we only do this processing once, thus this is still an approximation i.e. the resulting graph might have nodes with less than K links. 
    \item Using both locale and count filters, we finalize 6 versions of WebGraph dataset, summarized in Table \ref{tab:webgraph}.
\end{itemize}

\begin{table}
    \centering
    \begin{tabular}{|c|c|c|c|c|}
    \hline
      Dataset Name   & Top level domain & Min link count & Num nodes & Num edges\\ 
      \hline
      WebGraph-sparse & & 10  &365.4M& 29904M\\
      WebGraph-dense &  & 50 & 136.5M & 22158M \\
      WebGraph-de-sparse & de & 10 & 19.7M & 1192M\\
      WebGraph-de-dense & de & 50& 5.7M & 824M \\
      WebGraph-in-sparse & in & 10 & 1.5M & 149M \\
      WebGrpah-in-dense & in & 50 & 0.5M & 122M\\
      \hline
    \end{tabular}
    \caption{Stats for different version of WebGraph dataset.}
    \label{tab:webgraph}
\end{table}

We create and publish training and testing splits for evaluation purposes. The linkage graph is split by row (\emph{source link}) such that 90\% of the source links are in the training set and 10\% in the testing set. 
In the test set, for every source link, we hold out 25\% of the outlinks as ground truth and rest of the outlinks are meant to be used to obtain the embedding of the source link through Eq.~(\ref{eq:ialsupdate}). The embedding of the source link is then used to retrieve the top-K nearest neighbor predictions. These predictions are compared with the ground truth outlinks for computing evaluation metrics.
This splitting protocol is also known as strong generalization~\cite{marlin:thesis}.

\section{Experiments}\label{sec:experiments}
In this section, we describe our experiments on the WebGraph datasets. We further perform scaling analyses of our proposed implementation.

\subsection{Results}
We finally present our results for all the WebGraph variants. Based on our observations related to precision and choice of linear solvers: (1) we use bfloat16 to store the embedding tables and cast the input to the linear solvers to float32 and (2) Conjugate Gradients was employed for all the results we obtain since that was the fastest linear solver across the board. We use embeddings of dimension 128 and train the model for 16 epochs. In our experience, hyperparameter tuning over both norm penalty ($\lambda$) and unobserved weight ($\alpha$) has been indispensable for good results. On all the WebGraph datasets, we perform a grid search over 
\begin{itemize}
    \item $\lambda \in [\num{5e-2}, \num{1e-2}, \num{5e-3}, \num{1e-3}, \num{5e-4}, \num{1e-4}]$ 
    \item $\alpha \in [\num{1e-3}, \num{5e-4}, \num{1e-4}, \num{5e-5}, \num{1e-5}, \num{5e-6}, \num{1e-6}]$
\end{itemize}

Table \ref{tab:webgraph_perf} summarizes the best found hyperparameters and evaluation metrics for all variants of WebGraph.

\begin{table}
    \centering
    \begin{tabular}{|c|c|c|c|c|}
    \hline
      Dataset Name  & $\lambda$ & $\alpha$ & Recall@20 & Recall@50\\ 
      \hline
      WebGraph-sparse & \num{5e-2} & \num{1e-6}  & $0.365^{*}$& $0.377^{*}$\\
      WebGraph-dense & \num{1e-2} & \num{1e-5} & $0.652^{*}$ & $0.724^{*}$\\
      WebGraph-de-sparse & \num{1e-2} & \num{5e-6} & 0.901 & 0.936\\
      WebGraph-de-dense & \num{5e-4} & \num{5e-5} & 0.946 & 0.964\\
      WebGraph-in-sparse & \num{5e-3} & \num{1e-4} & 0.909 & 0.941\\
      WebGraph-in-dense & \num{1e-3} & \num{1e-3} & 0.965 & 0.974\\
      \hline
    \end{tabular}
    \caption{Training stats for different version of WebGraph dataset. All the models were trained for 16 epochs with embedding size of 128. Note that for WebGraph-dense and WebGraph-sparse, the recall numbers were calculated using approximate top K since exact top K would be too slow for this size. Thus the recall numbers are only an estimate of the true recall. They should be a lower bound of true recall numbers with high probability.}
    \label{tab:webgraph_perf}
\end{table}

As shown in Table \ref{tab:webgraph_perf}, we measure Recall@20 and Recall@50 on the test set for all the WebGraph variants. One striking thing to note is that iALS (Algorithm \ref{alg:ials}) is very effective for smaller variants of WebGraph like WebGraph-in and 
WebGraph-de. Moreover, the results indicate that iALS makes use of the additional linkage data and is more effective on denser versions of WebGraph across the board. 

In order to understand the structure of the WebGraph data that iALS is able to extract, we also spot check predictions from the model and perform a more qualitative analysis. One hypothesis for high recall numbers is that the dataset is riddled with outlinks to popular social media web links (facebook, twitter) and iALS is picking up on that. If this is indeed true, it makes the dataset less valuable and our results less impressive since a simple popularity based recommender which recommends these popular links all the time would also perform quite well. In reality, our observations from the predictions of the trained models suggest that iALS is able to learn to put web links from the same domain name nearby in the embedding space. We include a few sample predictions in Appendix \ref{sec:examples}.

\subsection{Scaling Analysis}
\label{subsec:empirical_scaling}
\begin{figure}
     \centering
     \begin{subfigure}[b]{0.4\textwidth}
         \centering
         \includegraphics[width=\textwidth]{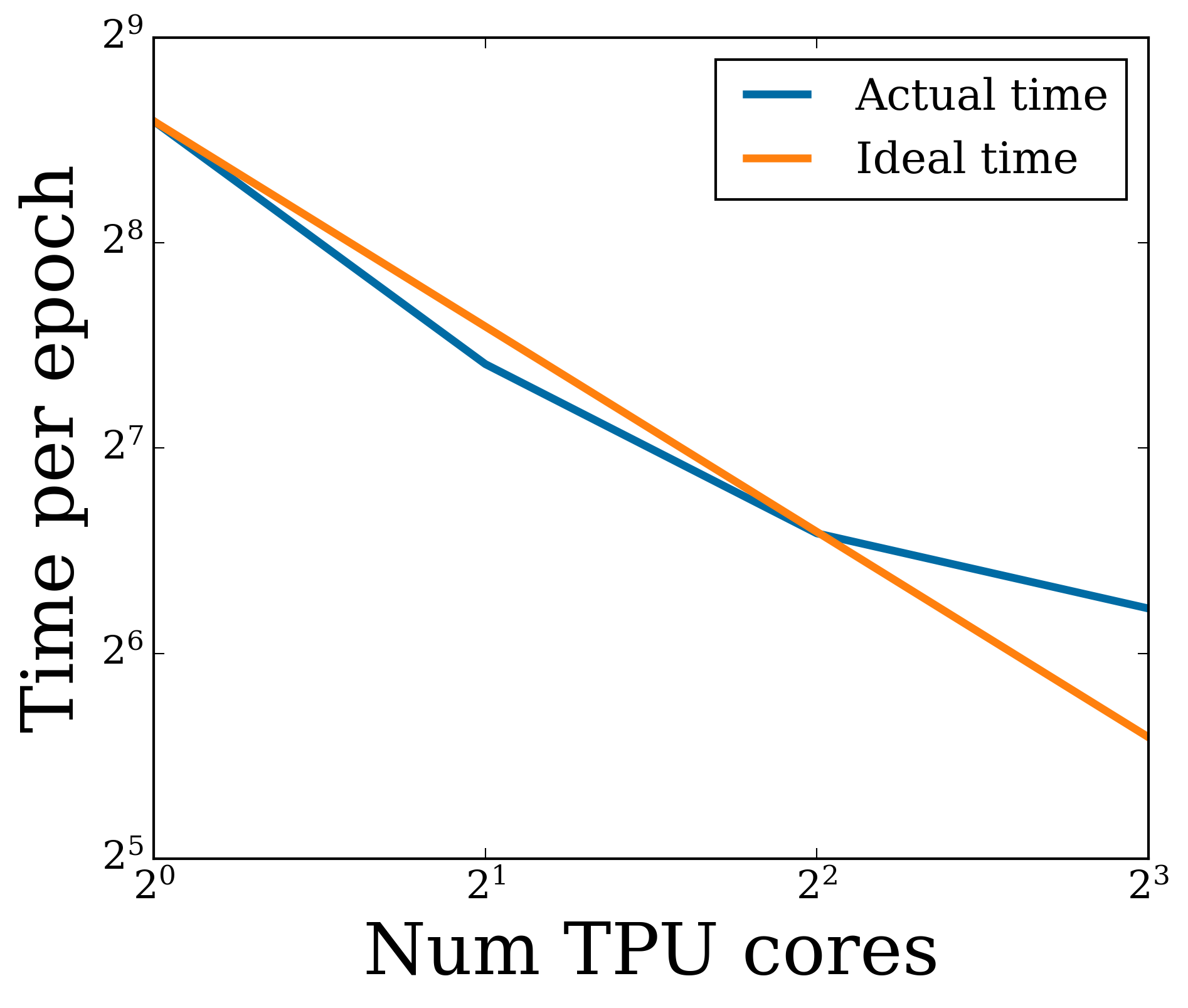}
         \caption{WebGraph-de-dense}
         \label{subfig:scale_de_dense}
     \end{subfigure}
     \hspace{0.07\textwidth}
     \begin{subfigure}[b]{0.4\textwidth}
         \centering
         \includegraphics[width=\textwidth]{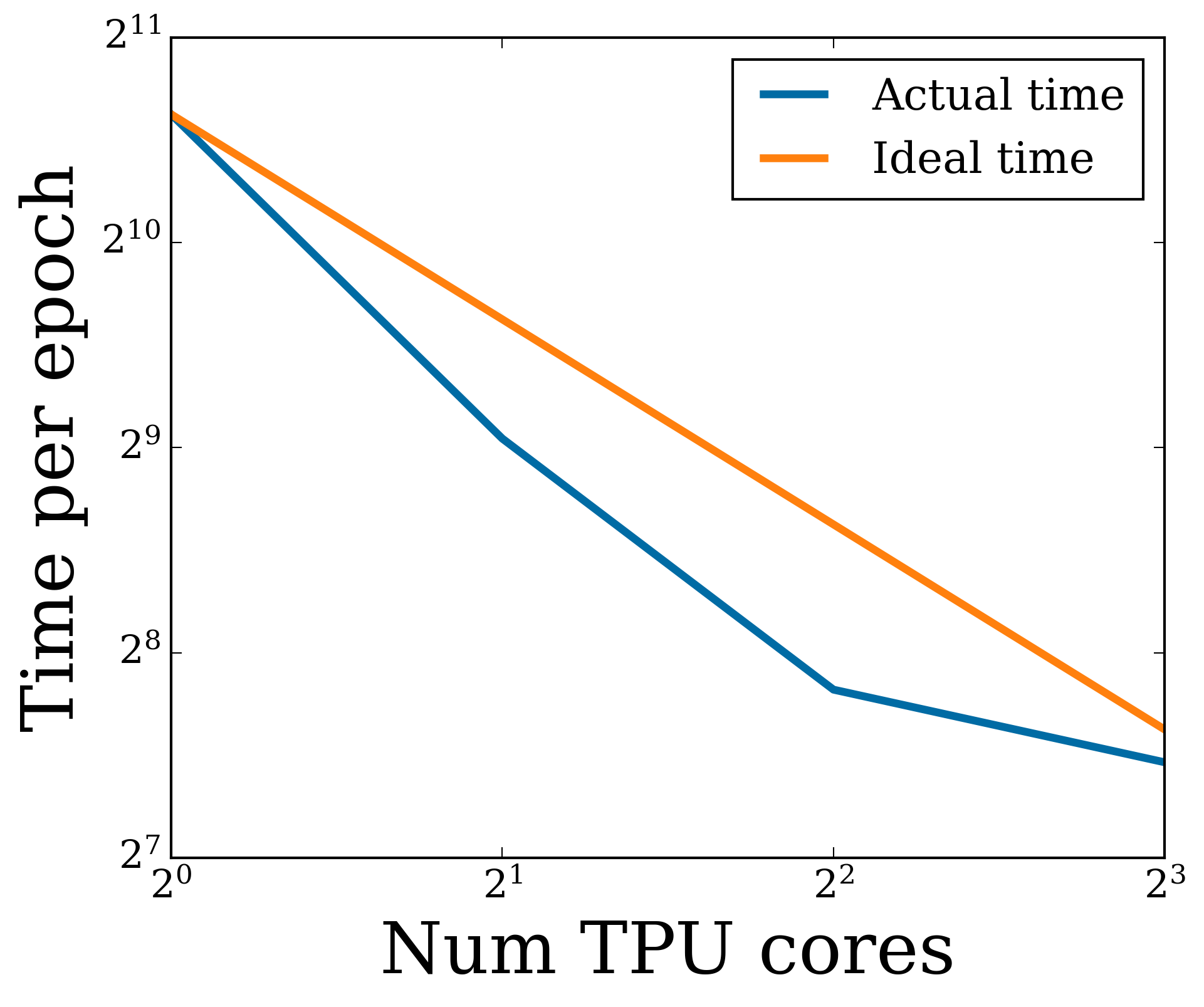}
         \caption{WebGraph-de-sparse}
         \label{subfig:scale_de_sparse}
     \end{subfigure}
    \begin{subfigure}[b]{0.4\textwidth}
         \centering
         \includegraphics[width=\textwidth]{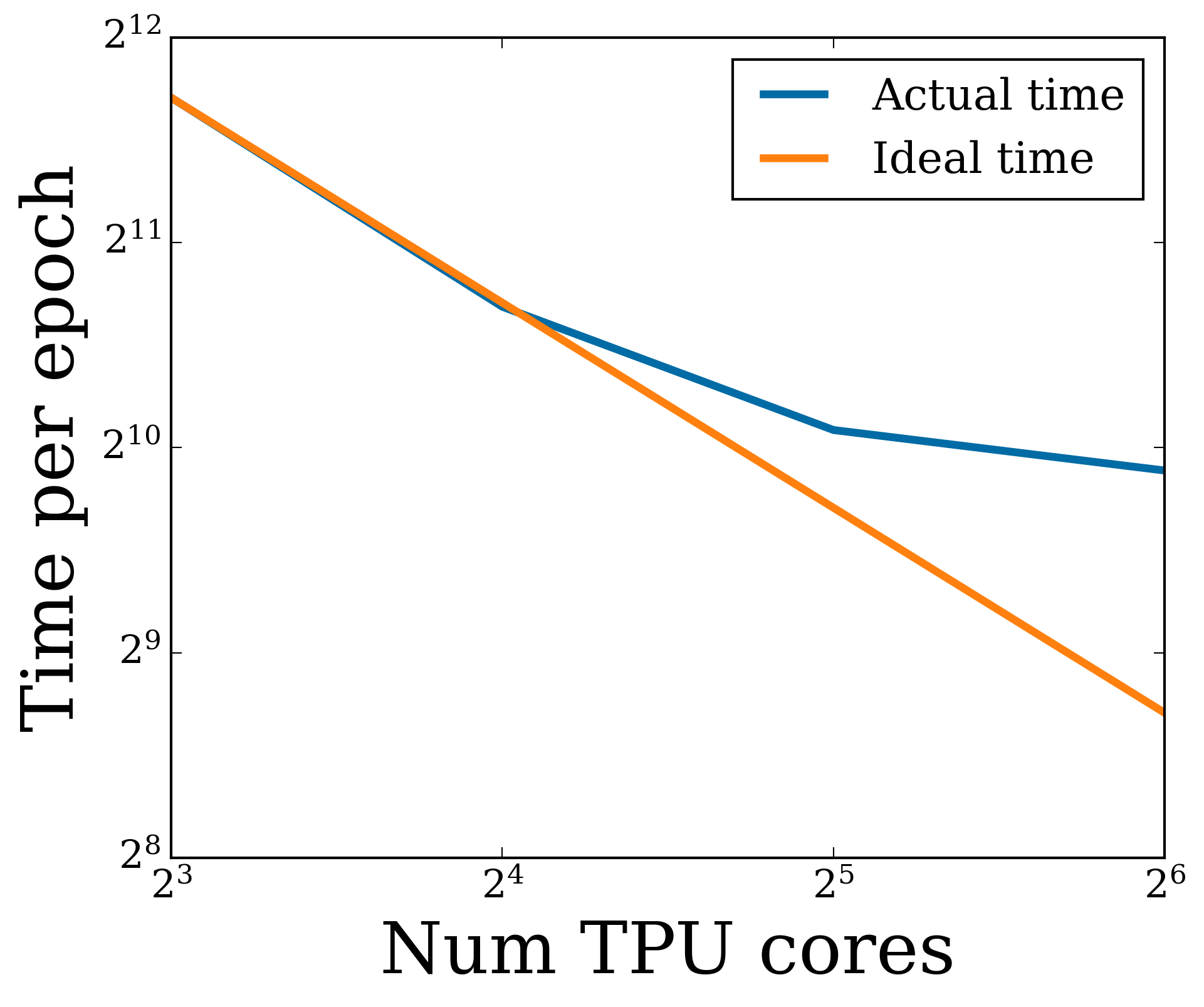}
         \caption{WebGraph-dense}
         \label{subfig:scale_dense}
     \end{subfigure}
     \hspace{0.07\textwidth}
     \begin{subfigure}[b]{0.4\textwidth}
         \centering
         \includegraphics[width=\textwidth]{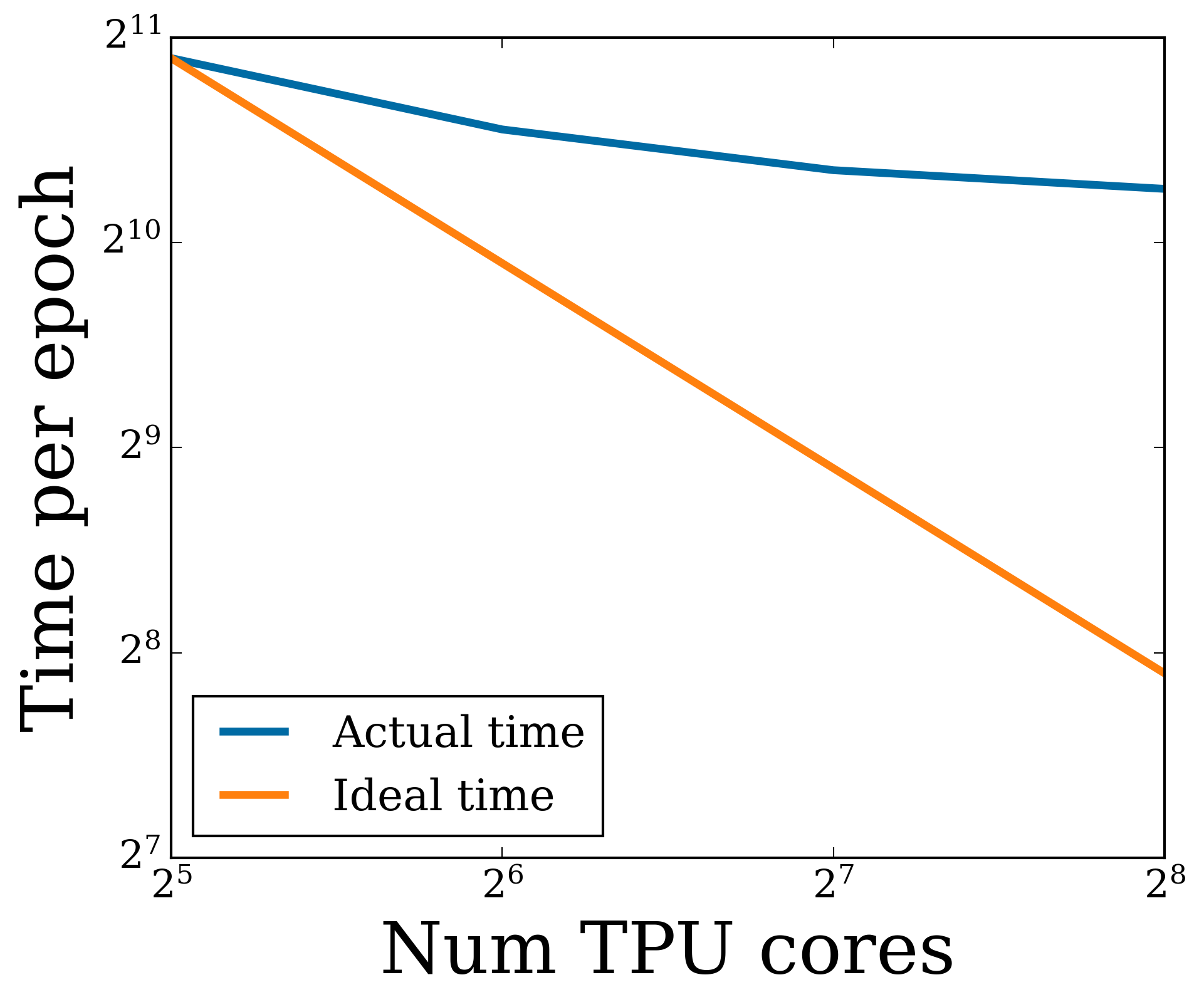}
         \caption{WebGraph-sparse}
        \label{subfig:scale_sparse}
     \end{subfigure}
     \hfill
        \caption{Scaling analysis of running time as number of TPU cores are increased. Each figure plots the time taken to train for one epoch in seconds.}
        \label{fig:scaling}
\end{figure}
In this Section, we analyze the scaling properties of ALX as a function of both dataset size and number of TPU available. Since we wanted to maximize the size of the input sparse matrix our system could handle, ALX shards the embedding tables uniformly across TPU cores. This also means that naively scaling the number of available TPU cores to an arbitrary large number can lead to need for excessive network communication. Thus, as we increase the number of TPU cores, we should be able to see a linear decrease in training time up to a certain point, after which the network overhead would prevent the training time to go down in a linear fashion.

In order to confirm our hypothesis, we analyze scaling properties of 4 biggest WebGraph variants in terms of training time as we increase the number of available TPU cores. As shown in Figure \ref{fig:scaling}, even empirically, we do observe the predicted linear decrease in training time up to a sweet spot, after which the network overhead start becoming the bottleneck.

Another thing to note is that, each TPU v3 core has a limited memory (16 GiB). For smaller datasets, whole embedding tables can be fit in a single core. But for larger problems, we need a minimum number of TPUs just to store the embedding tables. For example WebGraph-dense and WebGraph-sparse, as shown in Figure \ref{fig:scaling}, needs at least 8 and 32 TPUs respectively in order to even begin training. For WebGraph-dense we see linear decrease in training time up to 16 TPU cores. For WebGraph-sparse, initial requirement is large enough for it to be bottleneck-ed by network overhead even with a minimum number of TPUs required to store the embedding tables in memory.

\section{Conclusion}\label{sec:conclusion}
We designed, implemented and analyzed a new TPU based architecture for large scale distributed matrix factorization using Alternating Least Squares. The design circumvents specific restrictions from XLA (e.g. static shapes) while exploiting favorable properties of the TPU architecture like dedicated chip-to-chip inter-connects and move from exact solvers (like Cholesky) to iterative solvers like Conjugate Gradients.

In addition to the ALX framework, we also release a large scale link prediction dataset called WebGraph, based on data we obtained from CommonCrawl. Our aim for releasing these datasets is to spur research into developing techniques for handling very large scale sparse matrices. We also use this dataset to illustrate the speed and scaling abilities of ALX.

The code will be open-sourced and can be easily run on Google Cloud. In fact, we illustrated that medium scale sparse matrix (WebGraph-dense of size 135M x 135M with 22B edges) can be factorized in a colab connected to 8 TPU cores in less than a day. Finally, we have designed the ALX framework with scalability in mind. With 256 TPU cores, one epoch of the largest WebGraph variant, WebGraph-sparse (365M x 365M sparse matrix) takes around 20 minutes to finish (5.5 hours for the whole training run). The final model has around 100B parameters. Based on the projections for bigger datasets and in our experience, ALX can comfortably scale to matrices of 1B x 1B in size. We hope that our work inspires further improvements in both scalable methods and implementations of large scale matrix factorization.

\section{Acknowledgment}\label{sec:ack}
We thank many Google colleagues for helping at various stages of this project. In particular, we are grateful to the JAX team for numerous discussions, especially James Bradbury and Skye Wanderman-Milne; Blake Hechtman for help with XLA and Rasmus Larsen for useful discussions about performance of linear solvers on TPUs. We also thank Nicolas Mayoraz and John Anderson for reading earlier drafts of this paper and providing valuable feedback.

{
\bibliography{refs}
\bibliographystyle{plain}
}


\clearpage
\appendix

\section{Examples}
\label{sec:examples}

Nearest neighbors predictions from the model trained with WebGraph-IN-dense and WebGraph-de-sparse data respectively.

\begin{figure*}
    \centering
\includegraphics[width=1.0\textwidth]{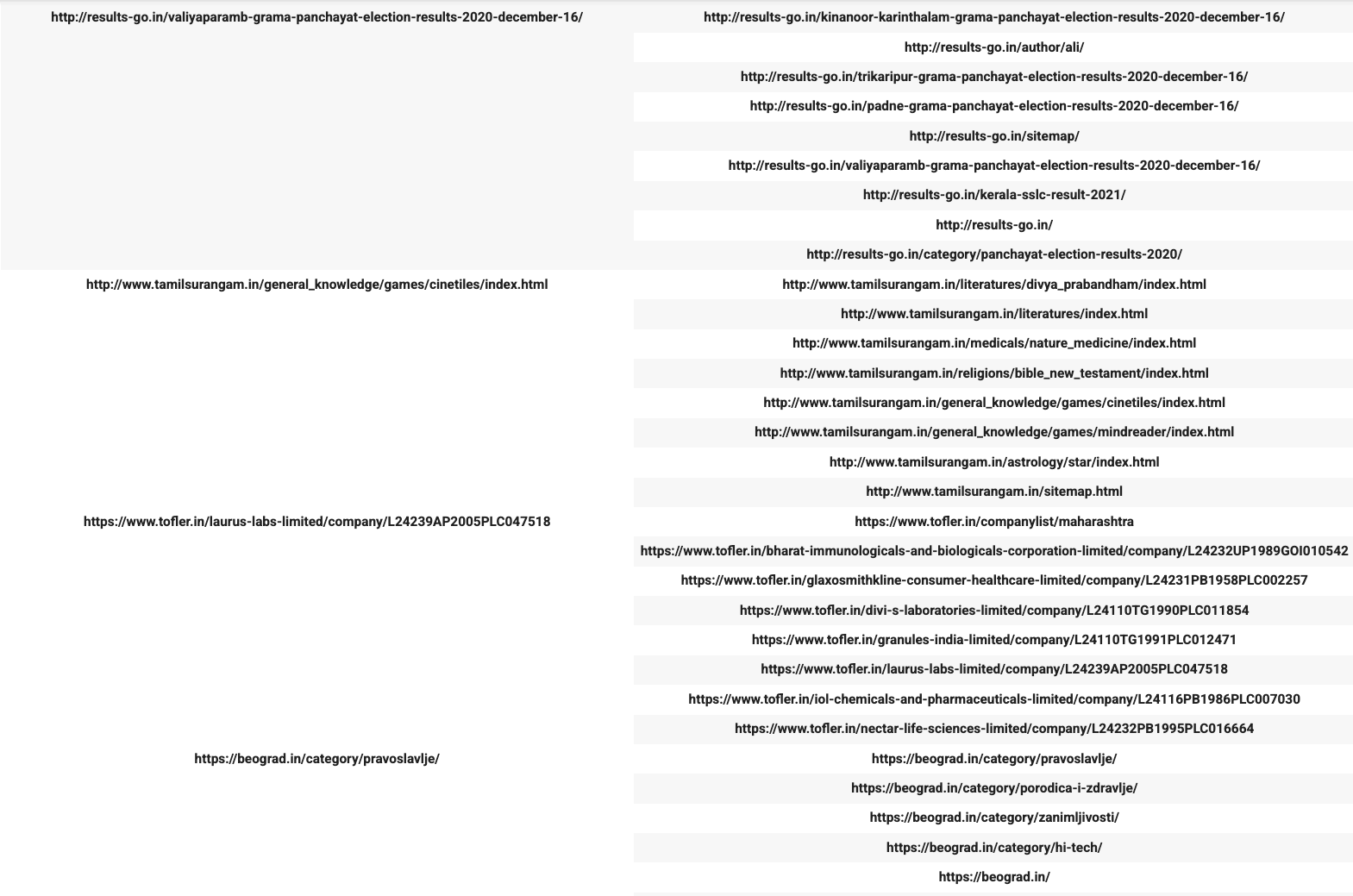}
    \caption{WebGraph-IN-dense example predictions}\label{fig:in_examples}
\end{figure*}

\begin{figure*}
    \centering
\includegraphics[width=1.0\textwidth]{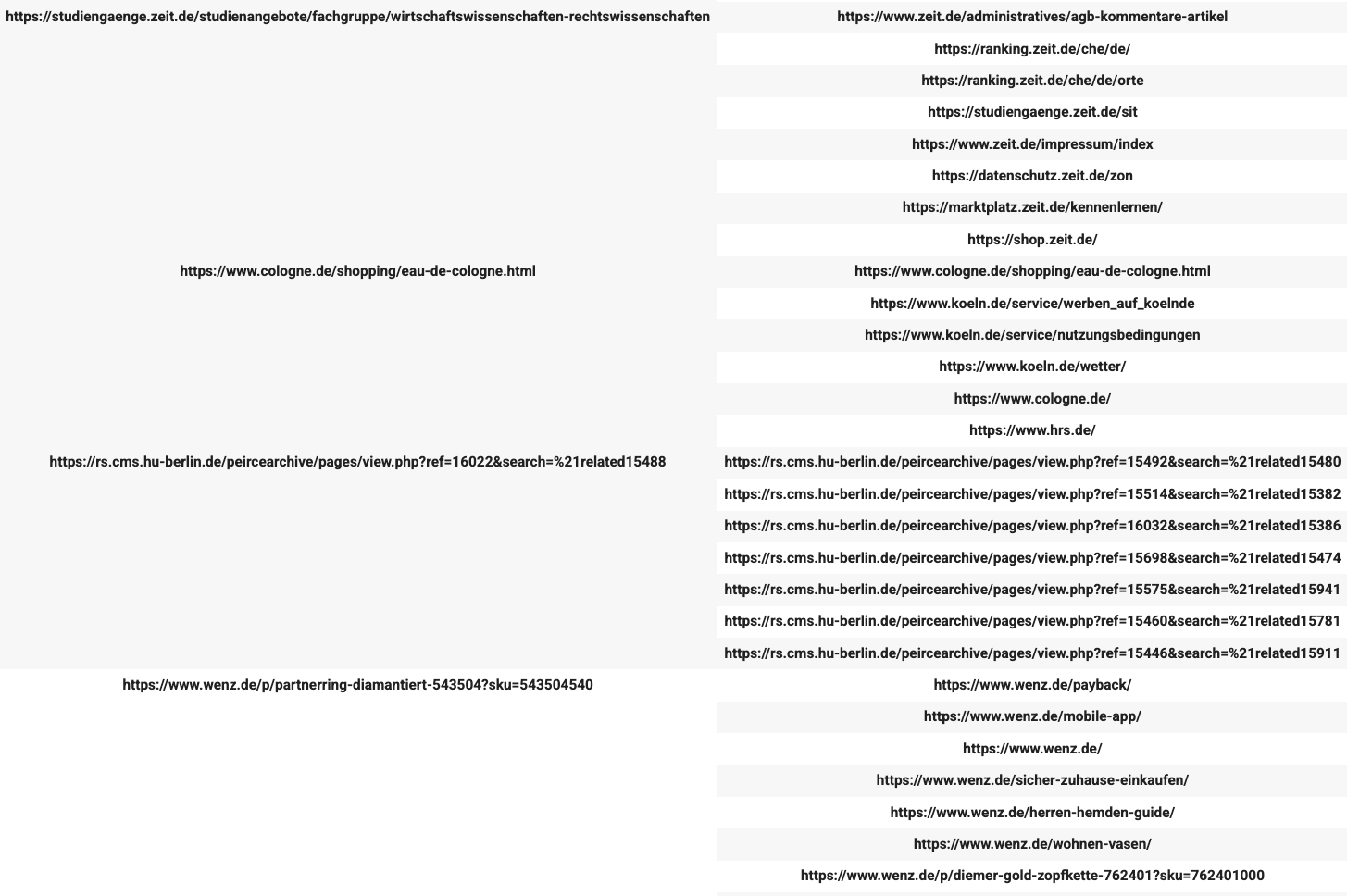}
    \caption{Webgraph-DE-sparse example predictions}\label{fig:de_examples}
\end{figure*}

\end{document}